\documentclass[10pt,twocolumn,letterpaper]{article}

\usepackage{cvpr}              % To produce the CAMERA-READY version
\definecolor{cvprblue}{rgb}{0.21,0.49,0.74}
\usepackage[pagebackref,breaklinks,colorlinks,allcolors=cvprblue]{hyperref}

\usepackage[T1]{fontenc}
\usepackage[utf8]{inputenc}
\usepackage{mathtools}

\usepackage{amssymb,mathrsfs}% Typical maths resource packages
\usepackage{amsthm}
\usepackage{bm}
\usepackage{scalerel}
\usepackage{nicefrac}
\usepackage{microtype} 
\usepackage[shortlabels]{enumitem}
\usepackage{graphicx}
\usepackage{epstopdf}
\DeclareGraphicsExtensions{.eps,.png,.jpg,.pdf}

\usepackage{url}
\usepackage{colortbl}
\usepackage{booktabs}
\usepackage{multirow}
\usepackage{colortbl,xcolor}
\usepackage{soul}
\usepackage{xparse,xstring}
\usepackage{calc}
\usepackage{etoolbox}

\makeatletter
\@ifpackageloaded{natbib}{
	\relax
}{
	\usepackage{cite}
}
\makeatother

\usepackage{array}
\newcolumntype{L}[1]{>{\raggedright\let\newline\\\arraybackslash\hspace{0pt}}m{#1}}
\newcolumntype{C}[1]{>{\centering\let\newline\\\arraybackslash\hspace{0pt}}m{#1}}
\newcolumntype{R}[1]{>{\raggedleft\let\newline\\\arraybackslash\hspace{0pt}}m{#1}}

\usepackage{glossaries}
\makeatletter
\sfcode`\.1006

\let\oldgls\gls
\let\oldglspl\glspl

\newcommand\fussy@ifnextchar[3]{%
	\let\reserved@d=#1%
	\def\reserved@a{#2}%
	\def\reserved@b{#3}%
	\futurelet\@let@token\fussy@ifnch}
\def\fussy@ifnch{%
	\ifx\@let@token\reserved@d
		\let\reserved@c\reserved@a
	\else
		\let\reserved@c\reserved@b
	\fi
	\reserved@c}

\renewcommand{\gls}[1]{%
\oldgls{#1}\fussy@ifnextchar.{\@checkperiod}{\@}}
\renewcommand{\glspl}[1]{%
\oldglspl{#1}\fussy@ifnextchar.{\@checkperiod}{\@}}

\newcommand{\@checkperiod}[1]{%
	\ifnum\sfcode`\.=\spacefactor\else#1\fi
}

\robustify{\gls}
\robustify{\glspl}
\makeatother

\newacronym{wrt}{w.r.t.}{with respect to}
\newacronym{RHS}{R.H.S.}{right-hand side}
\newacronym{LHS}{L.H.S.}{left-hand side}
\newacronym{iid}{i.i.d.}{independent and identically distributed}
\newacronym{SOTA}{SOTA}{state-of-the-art}
\usepackage{float}

\usepackage[capitalize]{cleveref}
\makeatletter
\def\cref@getref#1#2{%
  \expandafter\let\expandafter#2\csname r@#1@cref\endcsname%
  \expandafter\expandafter\expandafter\def%
    \expandafter\expandafter\expandafter#2%
    \expandafter\expandafter\expandafter{%
      \expandafter\@firstoffive#2}}% <-------- five
\def\cpageref@getref#1#2{%
  \expandafter\let\expandafter#2\csname r@#1@cref\endcsname%
  \expandafter\expandafter\expandafter\def%
    \expandafter\expandafter\expandafter#2%
    \expandafter\expandafter\expandafter{%
      \expandafter\@secondoffive#2}}% <----------- five
      
\AtBeginDocument{%
   \def\label@noarg#1{%
    \cref@old@label{#1}%
    \@bsphack%
    \edef\@tempa{{page}{\the\c@page}}%
    \setcounter{page}{1}%
    \edef\@tempb{\thepage}%
    \expandafter\setcounter\@tempa%
    \cref@constructprefix{page}{\cref@result}%
    \protected@write\@auxout{}%
      {\string\newlabel{#1@cref}{{\cref@currentlabel}%
      {[\@tempb][\arabic{page}][\cref@result]\thepage}{}{}{}}}% <----- five
    \@esphack}%
  \def\label@optarg[#1]#2{%
    \cref@old@label{#2}%
    \@bsphack%
    \edef\@tempa{{page}{\the\c@page}}%
    \setcounter{page}{1}%
    \edef\@tempb{\thepage}%
    \expandafter\setcounter\@tempa%
    \cref@constructprefix{page}{\cref@result}%
    \protected@edef\cref@currentlabel{%
      \expandafter\cref@override@label@type%
        \cref@currentlabel\@nil{#1}}%
    \protected@write\@auxout{}%
      {\string\newlabel{#2@cref}{{\cref@currentlabel}%
      {[\@tempb][\arabic{page}][\cref@result]\thepage}{}{}{}}}% <------- five
    \@esphack}%
    }  
\makeatother
\crefname{equation}{}{}
\Crefname{equation}{}{}
\crefname{claim}{claim}{claims}
\crefname{step}{step}{steps}
\crefname{line}{line}{lines}
\crefname{condition}{condition}{conditions}
\crefname{dmath}{}{}
\crefname{dseries}{}{}
\crefname{dgroup}{}{}
\crefname{page}{page}{pages}

\crefname{Problem}{Problem}{Problems}
\crefformat{Problem}{Problem~#2#1#3}
\crefrangeformat{Problem}{Problems~#3#1#4 to~#5#2#6}

\crefname{Theorem}{Theorem}{Theorems}
\crefname{Corollary}{Corollary}{Corollaries}
\crefname{Proposition}{Proposition}{Propositions}
\crefname{Lemma}{Lemma}{Lemmas}
\crefname{Definition}{Definition}{Definitions}
\crefname{Example}{Example}{Examples}
\crefname{Assumption}{Assumption}{Assumptions}
\crefname{Remark}{Remark}{Remarks}
\crefname{Rem}{Remark}{Remarks}
\crefname{remarks}{Remarks}{Remarks}
\crefname{Appendix}{Appendix}{Appendices}
\crefname{Supplement}{Supplement}{Supplements}
\crefname{Exercise}{Exercise}{Exercises}
\crefname{Theorem_A}{Theorem}{Theorems}
\crefname{Corollary_A}{Corollary}{Corollaries}
\crefname{Proposition_A}{Proposition}{Propositions}
\crefname{Lemma_A}{Lemma}{Lemmas}
\crefname{Definition_A}{Definition}{Definitions}

\usepackage{algorithm}
\usepackage{algpseudocode}

\ifx\loadbreqn\undefined
	\relax
\else
	\usepackage{breqn}
\fi

\makeatletter
\def\cleartheorem#1{%
    \expandafter\let\csname#1\endcsname\relax
    \expandafter\let\csname c@#1\endcsname\relax
}
\def\clearthms#1{ \@for\tname:=#1\do{\cleartheorem\tname} }
\makeatother

\ifx\renewtheorem\undefined
	\ifx\useTheoremCounter\undefined
		\newtheorem{Theorem}{Theorem}
		\newtheorem{Corollary}{Corollary}
		\newtheorem{Proposition}{Proposition}
		
	\else

	\fi

\fi

\theoremstyle{remark}

\theoremstyle{plain}

\newcommand{\qednew}{\nobreak \ifvmode \relax \else
		\ifdim\lastskip<1.5em \hskip-\lastskip
			\hskip1.5em plus0em minus0.5em \fi \nobreak
		\vrule height0.75em width0.5em depth0.25em\fi}



\NewDocumentCommand{\movedownsub}{e{^_}}{%
	\IfNoValueTF{#1}{%
		\IfNoValueF{#2}{^{}}% neither ^ nor _, do nothing; if no ^ but _, add ^{}
	}{%
		^{#1}% add superscript if present
	}%
	\IfNoValueF{#2}{_{#2}}% add subscript if present
}

\let\latexchi\chi
\RenewDocumentCommand{\chi}{}{\latexchi\movedownsub}

\newcommand{\calS}{\mathcal{S}}

\DeclareSymbolFont{bsfletters}{OT1}{cmss}{bx}{n}
\DeclareSymbolFont{ssfletters}{OT1}{cmss}{m}{n}
\DeclareMathSymbol{\bsfGamma}{0}{bsfletters}{'000}
\DeclareMathSymbol{\ssfGamma}{0}{ssfletters}{'000}
\DeclareMathSymbol{\bsfDelta}{0}{bsfletters}{'001}
\DeclareMathSymbol{\ssfDelta}{0}{ssfletters}{'001}
\DeclareMathSymbol{\bsfTheta}{0}{bsfletters}{'002}
\DeclareMathSymbol{\ssfTheta}{0}{ssfletters}{'002}
\DeclareMathSymbol{\bsfLambda}{0}{bsfletters}{'003}
\DeclareMathSymbol{\ssfLambda}{0}{ssfletters}{'003}
\DeclareMathSymbol{\bsfXi}{0}{bsfletters}{'004}
\DeclareMathSymbol{\ssfXi}{0}{ssfletters}{'004}
\DeclareMathSymbol{\bsfPi}{0}{bsfletters}{'005}
\DeclareMathSymbol{\ssfPi}{0}{ssfletters}{'005}
\DeclareMathSymbol{\bsfSigma}{0}{bsfletters}{'006}
\DeclareMathSymbol{\ssfSigma}{0}{ssfletters}{'006}
\DeclareMathSymbol{\bsfUpsilon}{0}{bsfletters}{'007}
\DeclareMathSymbol{\ssfUpsilon}{0}{ssfletters}{'007}
\DeclareMathSymbol{\bsfPhi}{0}{bsfletters}{'010}
\DeclareMathSymbol{\ssfPhi}{0}{ssfletters}{'010}
\DeclareMathSymbol{\bsfPsi}{0}{bsfletters}{'011}
\DeclareMathSymbol{\ssfPsi}{0}{ssfletters}{'011}
\DeclareMathSymbol{\bsfOmega}{0}{bsfletters}{'012}
\DeclareMathSymbol{\ssfOmega}{0}{ssfletters}{'012}

\makeatletter
\newcommand*\rel@kern[1]{\kern#1\dimexpr\macc@kerna}
\newcommand*\widebar[1]{%
  \begingroup
  \def\mathaccent##1##2{%
    \rel@kern{0.8}%
    \overline{\rel@kern{-0.8}\macc@nucleus\rel@kern{0.2}}%
    \rel@kern{-0.2}%
  }%
  \macc@depth\@ne
  \let\math@bgroup\@empty \let\math@egroup\macc@set@skewchar
  \mathsurround\z@ \frozen@everymath{\mathgroup\macc@group\relax}%
  \macc@set@skewchar\relax
  \let\mathaccentV\macc@nested@a
  \macc@nested@a\relax111{#1}%
  \endgroup
}
\makeatother

\DeclareMathOperator{\var}{var}

\DeclareMathOperator{\cov}{cov}

\newcommand{\ifbcdot}[1]{\ifblank{#1}{\cdot}{#1}}

\DeclarePairedDelimiterX\abs[1]{\lvert}{\rvert}{\ifbcdot{#1}}
\DeclarePairedDelimiterX\parens[1]{(}{)}{\ifbcdot{#1}}
\DeclarePairedDelimiterX\brk[1]{[}{]}{\ifbcdot{#1}}
\DeclarePairedDelimiterX\braces[1]{\{}{\}}{\ifbcdot{#1}}
\DeclarePairedDelimiterX\angles[1]{\langle}{\rangle}{\ifblank{#1}{\cdot,\cdot}{#1}}
\DeclarePairedDelimiterX\ip[2]{\langle}{\rangle}{\ifbcdot{#1},\ifbcdot{#2}}
\DeclarePairedDelimiterX\norm[1]{\lVert}{\rVert}{\ifbcdot{#1}}
\DeclarePairedDelimiterX\ceil[1]{\lceil}{\rceil}{\ifbcdot{#1}}
\DeclarePairedDelimiterX\floor[1]{\lfloor}{\rfloor}{\ifbcdot{#1}}

\DeclareFontFamily{U}{matha}{\hyphenchar\font45}
\DeclareFontShape{U}{matha}{m}{n}{
      <5> <6> <7> <8> <9> <10> gen * matha
      <10.95> matha10 <12> <14.4> <17.28> <20.74> <24.88> matha12
      }{}
\DeclareSymbolFont{matha}{U}{matha}{m}{n}
\DeclareFontSubstitution{U}{matha}{m}{n}

\DeclareFontFamily{U}{mathx}{\hyphenchar\font45}
\DeclareFontShape{U}{mathx}{m}{n}{
      <5> <6> <7> <8> <9> <10>
      <10.95> <12> <14.4> <17.28> <20.74> <24.88>
      mathx10
      }{}
\DeclareSymbolFont{mathx}{U}{mathx}{m}{n}
\DeclareFontSubstitution{U}{mathx}{m}{n}

\DeclareMathDelimiter{\vvvert}{0}{matha}{"7E}{mathx}{"17}
\DeclarePairedDelimiterX\vertiii[1]{\vvvert}{\vvvert}{\ifbcdot{#1}}

\DeclarePairedDelimiterXPP\trace[1]{\operatorname{Tr}}{(}{)}{}{\ifbcdot{#1}} % column vector
\DeclarePairedDelimiterXPP\col[1]{\operatorname{col}}{\{}{\}}{}{\ifbcdot{#1}} % column vector
\DeclarePairedDelimiterXPP\row[1]{\operatorname{row}}{\{}{\}}{}{\ifbcdot{#1}} % row vector
\DeclarePairedDelimiterXPP\erf[1]{\operatorname{erf}}{(}{)}{}{\ifbcdot{#1}}
\DeclarePairedDelimiterXPP\erfc[1]{\operatorname{erfc}}{(}{)}{}{\ifbcdot{#1}}
\DeclarePairedDelimiterXPP\KLD[2]{D}{(}{)}{}{\ifbcdot{#1}\, \delimsize\|\, \ifbcdot{#2}} % KL divergence
\DeclarePairedDelimiterXPP\op[2]{\operatorname{#1}}{(}{)}{}{#2} % general operator

\DeclarePairedDelimiterXPP\indicate[1]{{\bf 1}}{\{}{\}}{}{\ifbcdot{#1}}

\NewDocumentCommand\ofrac{s m}{%
	\IfBooleanTF#1%
	{\dfrac{1}{#2}}%
	{\frac{1}{#2}}%
}
\NewDocumentCommand\ddfrac{s m m}{%
	\IfBooleanTF#1%
	{\dfrac{\mathrm{d} {#2}}{\mathrm{d} {#3}}}%
	{\frac{\mathrm{d} {#2}}{\mathrm{d} {#3}}}%
}
\NewDocumentCommand\ppfrac{s m m}{%
	\IfBooleanTF#1%
	{\dfrac{\partial {#2}}{\partial {#3}}}%
	{\frac{\partial {#2}}{\partial {#3}}}%
}

\providecommand\given{}
\DeclarePairedDelimiterX\Set[2]\{\}{%
\renewcommand\given{\SetSymbol[\delimsize]{#1}}
#2
}
\DeclarePairedDelimiterX\Setc[1]\{\}{%
\renewcommand\given{\SetSymbol{:}}
#1
}

\NewDocumentCommand\set{s o m}{%
	\IfBooleanTF#1%
	{\IfValueTF{#2}{\Set*{#2}{#3}}{\Setc*{#3}}}%
	{\IfValueTF{#2}{\Set{#2}{#3}}{\Setc{#3}}}%
}

\NewDocumentCommand{\evalat}{ s O{\big} m e{_^} }{%
\IfBooleanTF{#1}%
{\left. #3 \right|}{#3#2|}%
\IfValueT{#4}{_{#4}}%
\IfValueT{#5}{^{#5}}%
}

\providecommand\given{}
\DeclarePairedDelimiterXPP\cprob[1]{}(){}{
\renewcommand\given{\nonscript\,\delimsize\vert\allowbreak\nonscript\,\mathopen{}}%
#1%
}
\DeclarePairedDelimiterXPP\cexp[1]{}[]{}{
\renewcommand\given{\nonscript\,\delimsize\vert\allowbreak\nonscript\,\mathopen{}}%
#1%
}

\DeclareDocumentCommand \P { s e{_^} d() g } {%
	\mathbb{P}%
	\IfBooleanTF{#1}%
		{
			\IfValueT{#2}{_{#2}}%
			\IfValueT{#3}{^{#3}}%
			\IfValueTF{#5}{\cprob{#4 \given #5}}{\IfValueT{#4}{\cprob{#4}}}%
		}%
		{
			\IfValueT{#2}{_{#2}}%
			\IfValueT{#3}{^{#3}}%
			\IfValueTF{#5}{\cprob*{#4 \given #5}}{\IfValueT{#4}{\cprob*{#4}}}%
		}%
}

\DeclareDocumentCommand \E { s e{_^} o g } {%
	\mathbb{E}%
	\IfBooleanTF{#1}%
		{
			\IfValueT{#2}{_{#2}}%
			\IfValueT{#3}{^{#3}}%
			\IfValueTF{#5}{\cexp{#4 \given #5}}{\IfValueT{#4}{\cexp{#4}}}%
		}%
		{
			\IfValueT{#2}{_{#2}}%
			\IfValueT{#3}{^{#3}}%	
			\IfValueTF{#5}{\cexp*{#4 \given #5}}{\IfValueT{#4}{\cexp*{#4}}}%		
		}%
}

\DeclareDocumentCommand \Var { s e{_^} d() g } {%
	\var%
	\IfBooleanTF{#1}%
		{
			\IfValueT{#2}{_{#2}}%
			\IfValueT{#3}{^{#3}}%
			\IfValueTF{#5}{\cprob{#4 \given #5}}{\IfValueT{#4}{\cprob{#4}}}%
		}%
		{
			\IfValueT{#2}{_{#2}}%
			\IfValueT{#3}{^{#3}}%	
			\IfValueTF{#5}{\cprob*{#4 \given #5}}{\IfValueT{#4}{\cprob*{#4}}}%		
		}%
}

\DeclareDocumentCommand \Cov { s e{_^} d() g } {%
	\cov%
	\IfBooleanTF{#1}%
		{
			\IfValueT{#2}{_{#2}}%
			\IfValueT{#3}{^{#3}}%
			\IfValueTF{#5}{\cprob{#4 \given #5}}{\IfValueT{#4}{\cprob{#4}}}%
		}%
		{
			\IfValueT{#2}{_{#2}}%
			\IfValueT{#3}{^{#3}}%	
			\IfValueTF{#5}{\cprob*{#4 \given #5}}{\IfValueT{#4}{\cprob*{#4}}}%		
		}%
}

\ExplSyntaxOn
\NewDocumentCommand \dist {m o o} {%
\mathrm{#1}\left(%
	\IfValueT{#3}{%
		\tl_if_blank:nTF{ #3 }{\cdot\, \middle|\, }{#3\, \middle|\, }%
	}
	\IfValueT{#2}{#2}%
\right)%
}
\ExplSyntaxOff

\NewDocumentCommand {\cbrace} {t+ D[]{black} D(){\widthof{#5}} m m } {%
	\begingroup%
		\color{#2}
		\IfBooleanTF{#1}{%
			\overbrace{#4}^%
		}{
			\underbrace{#4}_%
		}%
		{\parbox[c]{#3}{\centering\footnotesize{#5}}}%
	\endgroup% 
}

\let\oldforall\forall
\renewcommand{\forall}{\oldforall \, }

\let\oldexist\exists
\renewcommand{\exists}{\oldexist \, }

\makeatletter

\newcommand{\rankcolor}[2]{%
	\expandafter\renewcommand\csname #1\endcsname[1]{%
		\ifblank{##1}{%
			{\color{#2} \textbf{#2}}%
		}{%
			\ifmmode
				\textcolor{#2}{\bm{##1}}%
			\else%
				{\color{#2} \textbf{##1}}%
			\fi	
		}%
	}
}

\providecommand{\first}{}

\rankcolor{first}{red}
\rankcolor{second}{blue}
\rankcolor{third}{cyan}
\makeatother

\graphicspath{{./Figures/}{./figures/}}
\DeclareDocumentCommand{\includeCroppedPdf}{ o O{./Figures/} m }{
	\IfFileExists{#2#3-crop.pdf}{}{%
		\immediate\write18{pdfcrop #2#3.pdf #2#3-crop.pdf}}%
	\includegraphics[#1]{#2#3-crop.pdf}
}

\makeatletter
\newcommand*{\addFileDependency}[1]{% argument=file name and extension
  \typeout{(#1)}
  \@addtofilelist{#1}
  \IfFileExists{#1}{}{\typeout{No file #1.}}
}
\makeatother

\definecolor{gray90}{gray}{0.9}
\def\colorlist{red,blue,brown,cyan,darkgray,gray,lightgray,green,lime,magenta,olive,orange,pink,purple,teal,violet,white,yellow}

\makeatletter
\def\startcomment{[}
\ifx\nohighlights\undefined
	\newcommand{\createcolor}[1]{%
			\expandafter\newcommand\csname #1\endcsname[1]{{\color{#1} ##1}}%
	}
	\newcommand{\msout}[1]{\text{\color{green} \st{\ensuremath{#1}}}}
	\newcommand{\del}[1]{{\color{green}\ifmmode \msout{#1}\else\st{#1}\fi}}
\else
	\newcommand{\createcolor}[1]{%
			\expandafter\newcommand\csname #1\endcsname[1]{%
				\noexpandarg%
				\StrChar{##1}{1}[\firstletter]%
				\if\firstletter\startcomment%
					\relax
				\else%
					##1
				\fi
			}%
	}
	\newcommand{\msout}[1]{}
	\newcommand{\del}[1]{}
\fi

\def\@tempa#1,{%
    \ifx\relax#1\relax\else
        \createcolor{#1}%
        \expandafter\@tempa
    \fi
}
\expandafter\@tempa\colorlist,\relax,
\makeatother

\newcommand{\hhide}[1]{}
\ifx\diagnoselabel\undefined
	\relax
\else
	\makeatletter
	\def\@testdef #1#2#3{%
		\def\reserved@a{#3}\expandafter \ifx \csname #1@#2\endcsname
			\reserved@a  \else
			\typeout{^^Jlabel #2 changed:^^J%
				\meaning\reserved@a^^J%
				\expandafter\meaning\csname #1@#2\endcsname^^J}%
			\@tempswatrue \fi}
	\makeatother
\fi

\renewcommand{\first}[1]{\ifblank{#1}{\textbf{bold}}{\textbf{#1}}}

\newacronym{ODEs}{ODEs}{ordinary differential equations}

\renewcommand{\subsubsection}[1]{\noindent\textbf{#1}}

\usepackage{makecell}

\def\paperID{0000} % * Enter the Paper ID here
\def\confName{0000}
\def\confYear{2025}

\title{CodeBoost: Boosting Code LLMs by Squeezing Knowledge from Code Snippets with RL}

\author{
Sijie Wang\textsuperscript{*1} \quad 
Quanjiang Guo\textsuperscript{*2} \quad 
Kai Zhao\textsuperscript{1} \quad 
Yawei Zhang\textsuperscript{1} \quad 
Xin Li\textsuperscript{1} \quad 
Xiang Li\textsuperscript{1} \quad \\
Siqi Li\textsuperscript{1} \quad 
Rui She\textsuperscript{3} \quad 
Shangshu Yu\textsuperscript{4} \quad  
Wee Peng Tay\textsuperscript{1} \\ 
{\small \textsuperscript{1}Nanyang Technological University \quad }
{\small\textsuperscript{2}University of Electronic Science and Technology of China \quad }\\
{\small\textsuperscript{3}Beihang University \quad }
{\small\textsuperscript{4}Northeastern University, China }
\\ 
{\small
  \includegraphics[height=16pt]{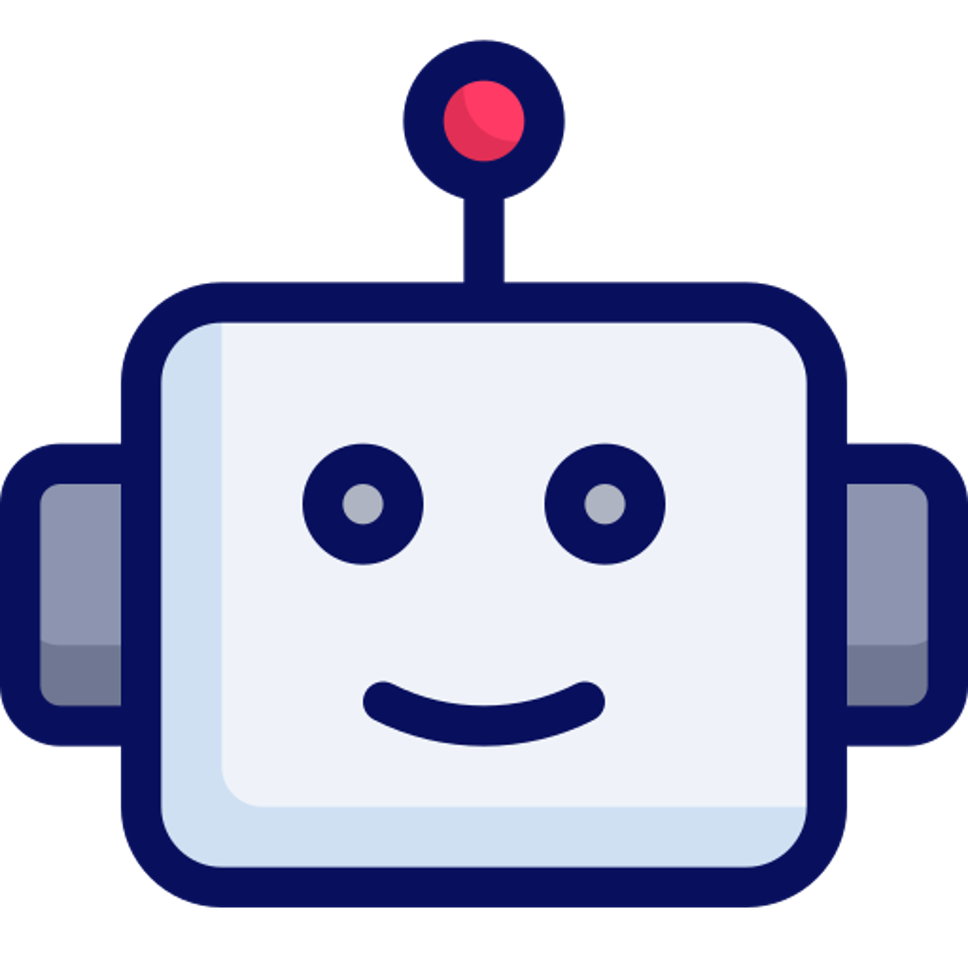}
  \url{https://github.com/sijieaaa/CodeBoost}
}
}

\begin{document}

% \maketitle

\twocolumn[{
\maketitle

\begin{center}
\setlength{\tabcolsep}{4pt}
\small
\vspace{-16pt}
\begin{tabular}{l|cc cc cc c | c  c c c}
\toprule
Model & \makecell{BCB (Hard)\\Complete} & \makecell{BCB (Hard)\\Instruct}  & \makecell{CRUXEval\\Output}  & \makecell{CRUXEval\\Input} & \makecell{MBPP} & \makecell{EvalPlus\\MBPP+} & \makecell{LiveCodeBench\\2501-2505} & \makecell{Total\\Perf.} \\
\midrule
Qwen2.5-Coder-7B-Instruct &  21.6 & 18.9 & 55.8 & 57.0 & 82.0 & 71.4 & 20.3 & 327.0\\
+ CodeBoost & \first{23.0} & \first{19.6} & \first{56.2} & \first{57.9} & \first{83.6} & \first{72.8} & \first{21.5} & \first{334.6} \\
\midrule
Llama-3.1-8B-Instruct & 14.2 & 13.5 & 38.1 & 38.2 &  67.5 & 55.3 & 15.2  & 242.0\\
+ CodeBoost & \first{14.9} & \first{16.9} & \first{40.1} & \first{38.4} & \first{71.4} & \first{59.0} &  \first{17.4} & \first{258.1} \\
\midrule
Seed-Coder-8B-Instruct & \first{30.4} & 27.0 & 60.9 & 58.1 & 85.2 & 71.2 & \first{23.4} & 356.2 \\
+ CodeBoost & \first{30.4} & \first{28.4} & \first{61.0} & \first{58.8} & \first{86.2} & \first{71.4} & \first{23.4}  &  \first{359.6} \\
\midrule
Yi-Coder-9B-Chat & 16.2 & \first{14.2} & 52.0 & 46.2 & \first{82.8} & 69.6 & 20.8 & 301.8\\
+ CodeBoost & \first{17.6} & 13.5 & \first{53.4 }& \first{48.0 } & \first{82.8} & \first{70.1} & \first{21.5} & \first{306.9} \\
\bottomrule
\end{tabular}
\captionof{table}{Performance comparisons on different benchmarks. After integrating with our CodeBoost, the total performance score improvements can be shown in all models. The higher scores are highlighted with bold fonts.}
\label{tab:main}
\end{center}
\vspace{-4pt}
}]

\begin{abstract}
Code large language models (LLMs) have become indispensable tools for building efficient and automated coding pipelines. Existing models are typically post-trained using reinforcement learning (RL) from general-purpose LLMs using "human instruction-final answer" pairs, where the instructions are usually from manual annotations. However, collecting high-quality coding instructions is both labor-intensive and difficult to scale. On the other hand, code snippets are abundantly available from various sources. This imbalance presents a major bottleneck in instruction-based post-training. We propose CodeBoost, a post-training framework that enhances code LLMs purely from code snippets, without relying on human-annotated instructions. CodeBoost introduces the following key components: (1) maximum-clique curation, which selects a representative and diverse training corpus from code; (2) bi-directional prediction, which enables the model to learn from both forward and backward prediction objectives; (3) error-aware prediction, which incorporates learning signals from both correct and incorrect outputs; (4) heterogeneous augmentation, which diversifies the training distribution to enrich code semantics; and (5) heterogeneous rewarding, which guides model learning through multiple reward types including format correctness and execution feedback from both successes and failures. Extensive experiments across several code LLMs and benchmarks verify that CodeBoost consistently improves performance, demonstrating its effectiveness as a scalable and effective training pipeline. 
% Our code is available at \url{https://github.com/sijieaaa/CodeBoost}
\renewcommand{\thefootnote}{\fnsymbol{footnote}}
% \footnotemark[1] 
\footnotetext[1]{Sijie Wang and Quanjiang Guo contribute equally. Corresponding author: Quanjiang Guo.}
\renewcommand{\thefootnote}{\arabic{footnote}} 
\end{abstract}
\glsresetall

\section{Introduction}

Code large language models (LLMs) have become essential tools for building efficient and effective research and development pipelines~\cite{roziere2023codellama,hui2024qwen2.5-coder,seed2025seed-coder,young2024yi,dubey2024llama3,huang2024opencoder}. By interpreting input prompts, code LLMs can perform a wide range of coding tasks, such as code generation and completion.

LLM training generally consists of two stages: pre-training and post-training. In the pre-training stage~\cite{devlin2019bert,radford2018improving_gpt}, models learn general language representations and knowledge from large-scale corpora in a self-supervised manner. The post-training stage further aligns LLMs with human preferences~\cite{bai2022training_RLHF} or adapts them to specific tasks~\cite{shao2024deepseekmath,qwen2.5,yang2024swe-agent,xin2024deepseek-prover,hurst2024gpt-4o}, thereby improving usability and task-specific performance.

Code LLMs are typically post-trained from general-purpose LLMs to better address code-related objectives. Two main post-training approaches are widely used: supervised fine-tuning (SFT) and reinforcement learning (RL). SFT trains models on curated "human instruction–full answer" pairs, enabling them to generate complete responses based on human instructions. RL-based methods, in contrast, supervise only the final answer in "human instruction–final answer" pairs, using reward signals such as code execution correctness and response formatting. Compared to SFT, RL approaches allow code LLMs to explore beyond supervised data and discover more optimal and general solutions~\cite{guo2025deepseek-r1}.

However, collecting such instruction–answer pairs is both tedious and labor-intensive. Of the two components, coding instructions, such as questions, comments, or annotations, are particularly scarce and difficult to obtain, as they typically require manual creation by experts. In contrast, raw code is abundantly available from open-source platforms and public repositories.
This imbalance in data availability creates a major bottleneck for instruction-based post-training. It raises a natural question: can we further enhance code LLMs by leveraging the vast availability of raw code alone? This motivates the development of alternative training strategies that bypass human-annotated instructions and instead utilize code snippets to generate pseudo-instructions directly.

In response, we propose \textbf{CodeBoost}, a novel training pipeline designed to enhance code LLMs using only raw code snippets. 
First, we introduce maximum-clique curation to construct a diverse and representative training corpus, improving the effectiveness of model learning. 
Next, we employ bi-directional tasking, enabling code LLMs to extract knowledge from both forward execution prediction and backward code completion. 
We further incorporate error-aware prediction, allowing the model to learn from both successful and failed executions. 
To enrich the training process, we apply heterogeneous augmentation, which diversifies code semantics and implicit knowledge. 
Finally, we present heterogeneous rewarding, providing fine-grained supervision signals to guide model optimization. 
Extensive experiments demonstrate that CodeBoost consistently improves the performance of existing code LLMs across multiple benchmarks, validating its effectiveness.

\section{Related Work}

In this section, we review existing related works, including training methods and code LLMs.

\subsection{RL Methods}
RL provides a principled framework for optimizing non-differentiable objectives through iterative interaction with an environment. RL algorithms are commonly divided into value-based and policy-based approaches.

Value-based methods aim to learn a value function that estimates the expected return for each action, with the policy derived by selecting actions that maximize this value. A classic example is Q-Learning~\cite{watkins1992q_learning}, which forms the foundation for Deep Q-Networks (DQN)~\cite{mnih2013playing_DQN}, where deep neural networks are used to approximate the action-value function in complex, high-dimensional spaces.

Policy-based methods, in contrast, directly optimize a parameterized policy. Notable algorithms in this category include Proximal Policy Optimization (PPO)~\cite{schulman2017proximal_ppo}, Trust Region Policy Optimization (TRPO)~\cite{schulman2015trust_trpo}, Direct Preference Optimization (DPO)~\cite{rafailov2023direct_dpo}, Group Relative Preference Optimization (GRPO)~\cite{shao2024deepseekmath}, and Dynamic Sampling Policy Optimization (DAPO)~\cite{yu2025dapo}. These methods are particularly effective for aligning models with diverse feedback signals, making them well-suited for coding tasks.

\subsection{Pre-Training}
Pre-training~\cite{devlin2019bert,radford2018improving_gpt,radford2019language_gpt2,brown2020language_gpt3} is the foundational stage for LLMs, enabling them to acquire broad linguistic, semantic, and structural knowledge from massive text corpora. During this phase, models are trained on diverse datasets, including web pages, articles, and code, using self-supervised objectives such as next-token prediction or masked language modeling.

This process allows LLMs to learn rich contextual representations, syntactic patterns, and factual world knowledge, which can be effectively transferred to a wide range of downstream tasks. The quality, scale, and diversity of the pre-training corpus are critical for determining the model’s generalization ability, robustness, and transfer performance. Consequently, pre-training has become a central paradigm in modern natural language processing and serves as the backbone for high-performing LLMs.

\subsection{Code LLM Post-Training}
Post-training for code LLMs predominantly follows two paradigms: SFT and RL.
SFT is the most common approach, where a pre-trained model is fine-tuned on high-quality "human instruction-full answer" pairs. Models like WizardCoder~\cite{luo2023wizardcoder}, Codex~\cite{chen2021evaluating_codex}, CodeT5~\cite{wang2021codet5}, Code Llama~\cite{roziere2023codellama}, StarCoder~\cite{li2023starcoder,lozhkov2024starcoder2}, OpenCoder~\cite{huang2024opencoder}, Llama-3.1~\cite{dubey2024llama3}, DeepSeek-Coder~\cite{guo2024deepseek-coder}, and Yi-Coder~\cite{young2024yi} exemplify this method, achieving strong results on coding benchmarks by training on datasets collected from competitive programming platforms and other sources. The primary drawback of SFT is its reliance on human-annotated instructions, which are expensive to create and inherently limited in scale and diversity compared to the vast amount of available raw code.

In contrast, RL provides a powerful alternative by enabling models to learn directly from environmental feedback, which in coding tasks is typically derived from code execution results. CodeRL~\cite{le2022coderl} pioneers to leverage RL for code generation. This approach has been further advanced by methods such as RLTF~\cite{liu2023rltf} and other execution-based reward frameworks~\cite{shojaee2023ppocoder}, which refine the use of test-driven signals for model optimization. Recently, RL-based code LLMs have gained significant traction, with models like DeepSeek-Coder-V2~\cite{zhu2024deepseek-coder-v2}, DeepSeek-R1~\cite{guo2025deepseek-r1}, Qwen2.5-Coder~\cite{hui2024qwen2.5-coder}, Qwen3~\cite{qwen3}, and Seed-Coder~\cite{seed2025seed-coder} demonstrating strong performance across a variety of coding benchmarks.

However, existing RL pipelines still require human-annotated instructions to prompt the training process, where such data collection would be tedious and labor-intensive.
Building on recent advances, our proposed CodeBoost framework enables RL-based post-training of code LLMs using only raw code snippets, eliminating the need for human-annotated instructions. Extensive experiments demonstrate that CodeBoost effectively enhances code LLM performance, establishing a scalable and instruction-free paradigm for future development in the LLM community.

\subsection{Closed-Source Code LLMs}
Advances in auto-regressive generation models have significantly accelerated the development of LLMs. Today, LLMs are increasingly deployed across a wide range of domains. Specifically in the coding field, several proprietary models have achieved state-of-the-art performance across diverse coding tasks. Notable examples include OpenAI’s GPT~\cite{achiam2023gpt-4,hurst2024gpt-4o,jaech2024openai-o1}, Anthropic’s Claude\footnote{\url{https://www.anthropic.com/news/claude-4}}, Google’s Gemini\footnote{https://deepmind.google/models/gemini/}, and xAI’s Grok\footnote{\url{https://x.ai/news/grok-4}}, all of which demonstrate strong capabilities in coding.

\section{CodeBoost Pipeline}
In this section, we provide a comprehensive overview of our proposed CodeBoost pipeline. We collect code snippets from open-source datasets, followed by rigorous filtering and maximum-clique-based curation to ensure diversity and reduce redundancy. We then formulate two complementary training tasks: forward execution output prediction and backward code completion. RL is employed to optimize the model, with heterogeneous reward signals guiding the training process for robust and effective code understanding.

\subsection{Dataset Collection}
The initial step in our training pipeline involves constructing a robust and diverse dataset.\footnote{This work focuses on Python, a widely-used code language that serves as a representative example for our pipeline.} We source code snippets from several open-source Python datasets, including OpenCoder~\cite{huang2024opencoder}, CodeForces-CoTs~\cite{penedo2025codeforces}, and Open-Thoughts-114k~\cite{guha2025openthoughts}, which collectively encompass broad semantics.
To ensure compatibility and reliability during training, we apply rigorous filtering to exclude code snippets that fail execution, are excessively short, or involve visualization-related components.

\begin{figure}[!tb]
\centering
\includegraphics[width=1\linewidth]{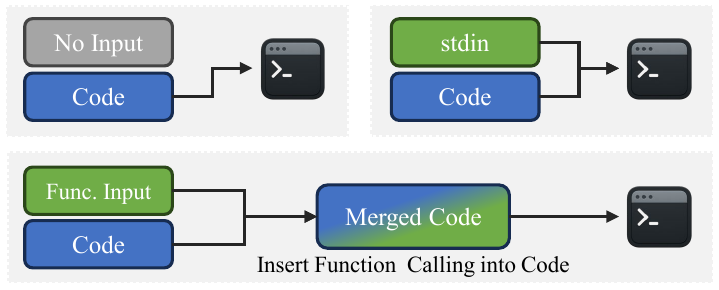}
\caption{Illustration on how different input types are handled in code execution.}
\label{fig:input_type}
\end{figure}

In addition to raw code snippets, these datasets also provide input examples, which are used for executing code. Python code snippets can generally be categorized based on input types: no-input code, standard-input (stdin) code, and function-input code. Execution strategies for these input types are illustrated in \cref{fig:input_type}. Specifically in our pipeline, no-input and stdin-based code can be executed directly without modification. In contrast, function-input code is preprocessed to insert input values explicitly into the code snippet before execution. This step ensures correctness and compatibility during execution. An overview of execution strategies for different input types is provided in \cref{fig:input_type}.

% In terms of execution, code can be grouped into directly executable code and function-only code. In terms of input type, there are no-input code, standard input (stdin) code, and function-input code.

\subsection{Maximum-Clique Dataset Curation}

\begin{figure}[!tb]
\centering
\includegraphics[width=1.0\linewidth]{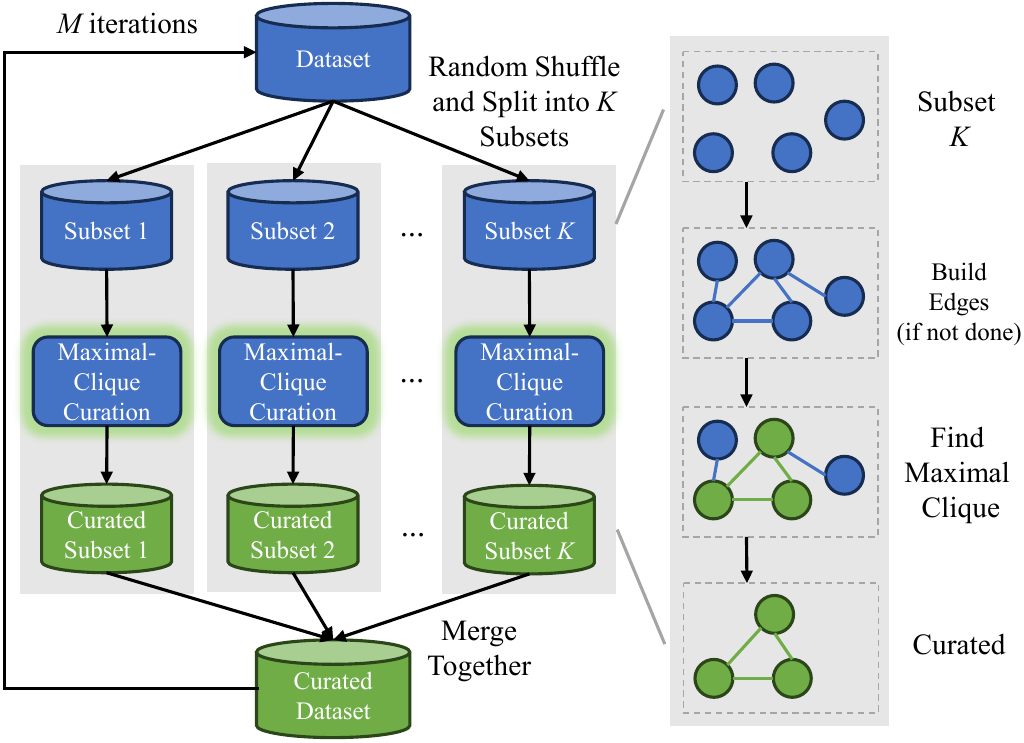}
\caption{Illustration of the maximum-clique curation pipeline.}
\label{fig:maximum_clique}
\end{figure}

Training on the aforementioned simply filtered dataset yields unsatisfactory results. We attribute this to the presence of extensive duplications within the dataset, which limit the ability of code LLMs to learn diverse patterns and knowledge from different code snippets. To address this issue, we employ advanced de-duplication techniques to further curate the dataset.

A common approach to de-duplication involves embedding code snippets into a feature space using encoders~\cite{muennighoff2022mteb,zhang2025qwen3-embedding}. However, embedding-based methods often fall short for code snippets due to the unique characteristics of code as a formal language with strict syntactic rules and logical structures. These features make it challenging to effectively capture code semantics through embeddings alone.

To overcome these limitations, we utilize pairwise structural distance, which is based on the Jaccard score and operates independently of embeddings. Using this metric, we construct a graph where vertices represent code snippets, and edges are determined by their structural distance. We then extract the maximum clique from the graph, ensuring that the selected code samples are diverse and representative. This process enables effective dataset curation and mitigates redundancy.

Specifically, given two code strings $S_i,S_j$, we define the code structure distance $d$ as:
\begin{align}\label{codedistance}
d(S_i, S_j)=\abs*{ ( \calS_i \cup \calS_j ) \setminus ( \calS_i \cap \calS_j ) },
\end{align}
where 
\begin{align*}
\calS_i=f_\mathrm{split}(S_i),\ \calS_j=f_\mathrm{split}(S_j),
\end{align*}
and $f_{\mathrm{split}}$ is the function that splits a string into a set that contains line-level sub-strings.
We then further define an indicator function $f_{\mathrm{ind}}$ to determine whether two code strings should be considered as different:
\begin{align}\label{eq:indicator_edge}
f_{\mathrm{ind}}(S_i, S_j) = \begin{cases}
1, & \text{if } d(S_i, S_j) \geq \gamma \cdot \min\left( \abs{ \mathcal{S}_i }, \abs{ \mathcal{S}_j } \right), \\
0, & \text{otherwise},
\end{cases}
\end{align}
where $\gamma$ is the threshold factor. 
Given the uncurated code dataset $\mathcal{D}=\set{S_i}_{i=1}^N$ containing $N$ code snippet strings, we construct a graph $\mathcal{G}=(\mathcal{V}, \mathcal{E})$. The vertices $\mathcal{V}$ represent the code snippets, while the edges $\mathcal{E}$ are defined based on \cref{eq:indicator_edge}. Specifically, two code snippets are considered sufficiently distinct and connected by an edge if their structural distance exceeds the threshold.

We extract the maximum clique $\mathcal{G}_\mathrm{clq}$ from $\mathcal{G}$. 
In graph theory, a maximum clique is defined as the largest complete subgraph within a graph, where every pair of vertices in the subset is connected by an edge. This ensures that all vertices in the clique are mutually adjacent, and no larger subset with this property exists. Consequently, in our extracted maximum clique $\mathcal{G}_\mathrm{clq}$, every pair of code snippets is sufficiently distinct, making the selected samples representative of the original dataset.

Extracting the full maximum clique from the entire dataset is computationally intensive in terms of both time and memory. To address this, we employ a divide-and-conquer strategy. The dataset is randomly divided into $K$ smaller subsets, and maximum clique extraction is performed independently on each subset, as shown in \cref{fig:maximum_clique}. 
The cliques with corresponding edges extracted from all subsets are then merged by taking their union, resulting in an intermediate curated dataset. In the next iteration, the intermediate dataset will be randomly divided into $K$ subsets again to proceed edge building and maximum-clique curation.
This process is repeated for $M$ iterations, progressively refining the selection. The final curated dataset serves as an approximate maximum clique of the original graph, reducing computational overhead while maintaining diversity and representativeness.

\subsection{RL Training}
RL enables LLMs to tackle diverse tasks by optimizing reward signals, reducing reliance on fully annotated ground truth (GT) answers typically required in SFT. Unlike SFT, RL provides models with greater flexibility to explore and discover optimal solutions~\cite{guo2025deepseek-r1,shao2024deepseekmath}, significantly enhancing their generalization capabilities. In domains such as mathematics and coding, RL is particularly effective, as GT answers are often deterministic, making correctness straightforward to quantify through rewards. In our pipeline, we adopt GRPO~\cite{guo2025deepseek-r1,shao2024deepseekmath}, a proven RL framework for LLM training. To ensure effective RL training, it is crucial to define versatile and well-suited tasks. To this end, we introduce our heterogeneous tasking strategy, detailed below.

\subsubsection{Bi-Directional Prediction.}
To enable effective training using raw code snippets, the tasks must be designed to be self-contained and complementary. 
One task is the forward task, which involves predicting the execution outputs of code snippets. In this task, code LLMs are required to analyze the entire structure of the code and reason step-by-step to accurately predict the final outputs. This task emphasizes comprehensive understanding and logical reasoning.

A natural second task is the backward task, which focuses on code completion. Here, code LLMs must fill in missing parts of incomplete code snippets to ensure that the completed code produces the expected execution outputs. This task requires the model to deeply understand code semantics and context while generating syntactically and functionally correct code.

\begin{figure}[!tb]
\centering
\includegraphics[width=1.01\linewidth]{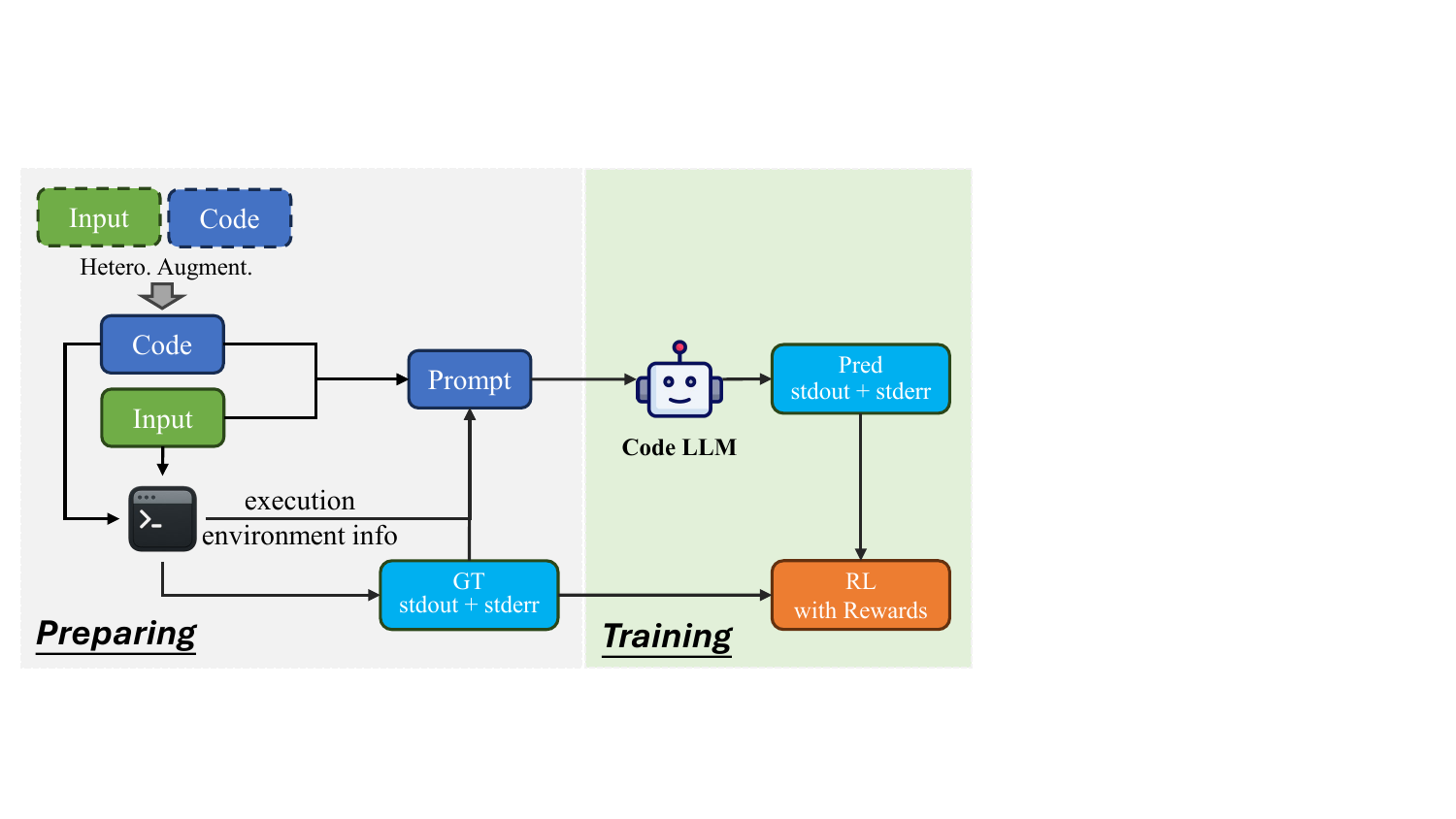}
\caption{Illustration of the forward task, where code LLMs are required to predict code execution outputs. }
\label{fig:forward}
\end{figure}

\begin{figure*}[!tb]
\centering
\includegraphics[width=0.9\linewidth]{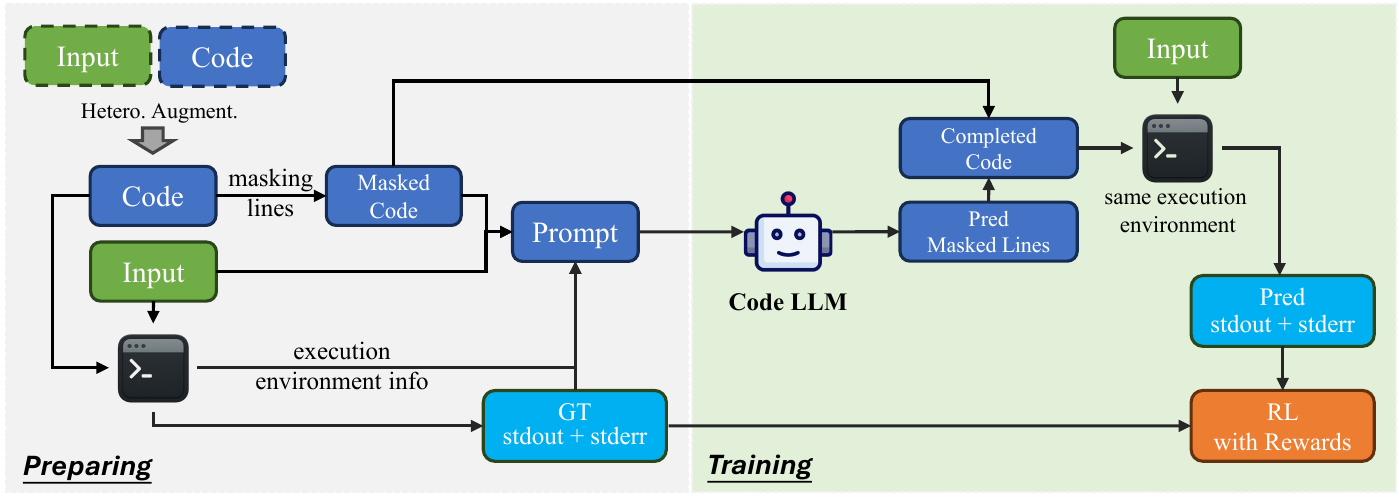}
\caption{Illustration of the backward task. Different from the forward task, in the backward task, code LLMs are required to predict the masked lines of the code, which are subsequently combined with the masked code to give predicted outputs.}
\label{fig:backward}
\end{figure*}

Together, these tasks form the bi-directional prediction framework, combining forward and backward reasoning to holistically enhance the coding capabilities of LLMs. An overview of these tasks is illustrated in \cref{fig:forward,fig:backward}.

\subsubsection{Error-Aware Prediction.}\label{sec:error_aware_prediction}
Existing code RL training predominantly focuses on error-free code snippets, thereby disregarding potentially valuable error signals. We argue that execution errors can carry rich supervision, particularly when they arise from deep, non-trivial logic. These latent errors emerge only when the model reasons with sufficient depth, offering a meaningful learning signal.

To harness this, we introduce error-aware prediction, a training paradigm that requires LLMs to jointly predict both standard output (stdout) and standard error (stderr), regardless of whether the code executes successfully. This setup presents a more challenging scenario, as the model cannot assume the absence of errors and must reason thoroughly to predict successes and failures.

\subsubsection{Heterogeneous Augmentation.}
To effectively extract knowledge from code snippets, it is essential to employ diverse augmentation strategies. A naive approach involves random augmentation of all characters in the code, which usually leads to syntax errors. 
Although these errors can enhance code LLMs’ syntax awareness, they prevent the models from engaging in deeper reasoning, thereby limiting their overall ability to understand complex code structures.

To address these limitations, we introduce heterogeneous augmentation, designed to facilitate deeper code comprehension. First, we apply digit-level augmentation specifically to isolated digits, as digits are frequently tied to critical aspects of code logic, such as arithmetic operations and indexing. Second, we implement logical-level augmentation by modifying operations (e.g., comparison, assignment, and unary operations) and conditions within the code. This is achieved by parsing the concrete syntax tree (CST) to ensure precise and meaningful perturbations. Logical-level augmentation alters the original code logic, presenting greater challenges for LLMs and encouraging deeper reasoning.

In the CodeBoost pipeline, we integrate heterogeneous augmentation for code snippets alongside digit augmentation for input values, ensuring a robust and diverse training process.

\subsubsection{Training Preparation.}
\begin{figure}[!htb]
\centering
\includegraphics[width=0.95\linewidth]{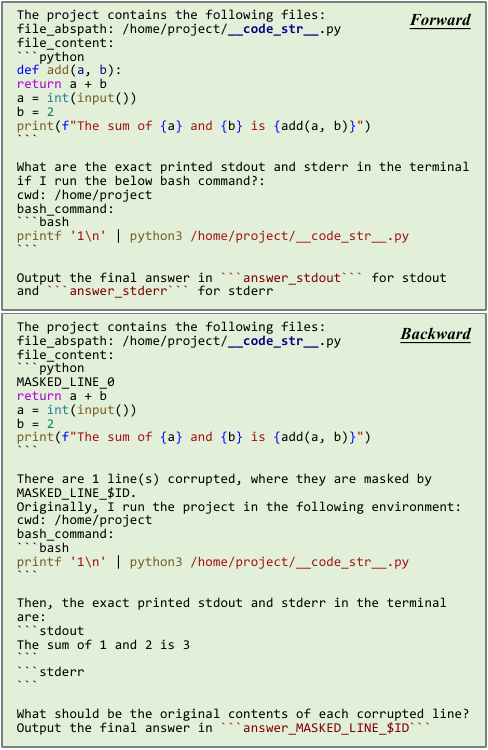}
\caption{The simplified prompt examples for both forward and backward tasks (with inputs). Code snippets are written in designated paths and executed by shell commands in the sandbox environment.}
\label{fig:prompt_simple}
\end{figure}

To achieve the aforementioned tasks, we need to build the GT execution outputs. 
As shown in \cref{fig:forward,fig:backward} for each training code snippet and corresponding input samples, we first apply heterogeneous augmentation onto them, which are subsequently executed by the code executor. As a result, the code executor will generate GT stdout, GT stderr, and environment information for this execution. This information is combined with code and input to form the prompt. We provide simplified prompt examples as shown in \cref{fig:prompt_simple} for both forward and backward tasks.

\subsection{Heterogeneous Rewarding}

In modern RL frameworks for code LLMs, the overall reward signal typically consists of two components: format reward and correctness reward. The format reward ensures that the model generates outputs in the correct structural format, enabling reliable parsing of the final answers. The correctness reward evaluates the functional accuracy of these parsed answers by comparing them against GT outputs. 

However, existing correctness reward designs are largely limited to cases where the GT code executes successfully, neglecting the potential learning value of execution error information. To address this limitation, we introduce heterogeneous rewarding, which integrates rewards for format adherence, stdout correctness, and stderr correctness, thereby providing a more comprehensive supervision signal.

\subsubsection{Rewarding the Forward Task.}
For the forward task, code LLMs are to predict the execution outputs, as shown in \cref{fig:forward}.
For each training sample, there exist predicted and GT stdout and stderr text strings. We denote them as $\mathrm{pred_{o}}$,  $\mathrm{pred_{e}}$, $\mathrm{GT_{o}}$, and $\mathrm{GT_{e}}$, respectively. We formulate the overall reward as:
\begin{align}\label{eq:reward_forward}
r=wr_\mathrm{format} + (1-w)(r_\mathrm{o} + \beta r_\mathrm{e})/2, \\
r_\mathrm{o}=\begin{cases}
1,        & \text{if } \mathrm{pred_o} = \mathrm{GT_o}, \\
0,        & \text{otherwise},
\end{cases} \\
r_\mathrm{e}=\frac{ \abs{ f_\mathrm{split}(\mathrm{pred_e}) \cap f_\mathrm{split}(\mathrm{GT_e}) } } { \abs{ f_\mathrm{split}(\mathrm{pred_e}) \cup f_\mathrm{split}(\mathrm{GT_e}) }  },
\end{align}
where $w,\beta$  are weight factors, and $r_\mathrm{format}$ is determined by whether the LLM's generated text follows the defined formats.
% More details about $r_\mathrm{format}$ be found in the supplement. 
For example, as in \cref{fig:prompt_simple}, the formats for forward tasks require markdown blocks with "answer\_stdout" and "answer\_stderr" tags, while backward tasks require "answer\_MASKED\_LINE\_\$ID" tags.
Note that, different from the stdout reward $r_\mathrm{o}$ that is defined in a hard form, the stderr reward $r_\mathrm{e}$ is defined in a soft form. This is because stderr information is more closely tied to the system and environment, making it difficult to predict exactly. Therefore, we define it in the Jaccard score style.

\subsubsection{Rewarding the Backward Task.}
In the backward task, the code will be randomly masked at the line level. Code LLMs are to predict the masked lines from the masked code snippets, as shown in \cref{fig:backward}. In this task, $\mathrm{pred_o}$ and $\mathrm{pred_e}$ are obtained by executing the completed code under the same execution environment, which is different from the process in the forward task. The reward formulation for the backward task follows the same structure as defined in \cref{eq:reward_forward}.

\section{Experiments}
\subsubsection{Implementation Details.}
Our curated dataset is built from off-the-shelf coding datasets. After performing basic filtering and clique-based curation, the resulted dataset contains a total of 58k code snippets. Our pipeline is implemented based on EasyR1\footnote{\url{https://github.com/hiyouga/EasyR1}}, where GRPO is used as the RL training method. The training is conducted for 1 epoch.
Code execution is performed within our custom-implemented sandbox environment, where each code snippet is written to a file and executed via shell commands, as illustrated in \cref{fig:prompt_simple}.
We apply CodeBoost onto representative existing code LLMs, including Qwen2.5-Coder-7B-Instruct~\cite{hui2024qwen2.5-coder}, Llama-3.1-8B-Instruct~\cite{dubey2024llama3}, Seed-Coder-8B-Instruct~\cite{seed2025seed-coder}, and Yi-Coder-9B-Chat~\cite{young2024yi}. These baseline LLM weights are from the latest updates (as of 2025-August-01) on their respective HuggingFace repositories. 
The tested benchmarks include BigCodeBench~\cite{zhuo2024bigcodebench}, CRUXEval~\cite{gu2024cruxeval}, MBPP~\cite{austin2021program_MBPP}, EvalPlus-MBPP+~\cite{liu2023your_evalplus}, and LiveCodeBench~\cite{jain2024livecodebench}, which cover various coding scenarios.
 The full prompt templates are attached in the supplement. For hyper-parameters, we use $\gamma=1,w=0.1,\beta=0.5$. 
All experiments are conducted on 8 Tesla A100-80GB GPUs. Due to page limits, more details can be found in the supplement.

\subsubsection{Main Results.} 
% We first test our CodeBoost on public benchmarks. As shown in \cref{tab:main}, our CodeBoost can boost all baseline code LLMs (though there exists a minor drop for Seed-Coder), where the total performance scores increase significantly. For Llama-3.1, it even obtains around 16 total score gains. While for the largest model Yi-Coder, which consists of 9B parameters, CodeBoost can still further boost it by 5 points more, which demonstrates the effectiveness of our design.

We evaluate CodeBoost on a suite of public benchmarks to assess its effectiveness across diverse code LLMs. As shown in \cref{tab:main}, CodeBoost consistently improves overall performance on all tested models.
For Qwen2.5-Coder-7B-Instruct, CodeBoost improves the total score from 327.0 to 334.6, achieving gains across most benchmarks, with BigCodeBench Instruct remaining unchanged. 
For Llama-3.1-8B-Instruct, CodeBoost provides a substantial boost of nearly 16 points in total score, with notable improvements on MBPP and EvalPlus.
While Seed-Coder-8B-Instruct already exhibits strong baseline performance, CodeBoost further elevates its total score from 356.2 to 359.6, marking it as the best-performing model overall. 
Finally, even for the largest model in our evaluation, Yi-Coder-9B-Chat, CodeBoost yields a 5-point gain, demonstrating its scalability and general applicability across architectures and model sizes.
These results highlight the robustness of CodeBoost and its effectiveness as a scalable, instruction-free enhancement to existing code LLMs.

\subsubsection{Ablation Studies.}
We evaluate the effectiveness of our proposed design through an ablation study, by removing individual components from the full CodeBoost pipeline. As summarized in \cref{tab:ablation}, each module contributes positively to the overall performance, highlighting the importance of their integration. 

Among these, the heterogeneous augmentation module stands out as a necessary component. Its removal results in a significant performance drop, underscoring its role in enhancing model generalization through diverse code variations.

In the context of bi-directional tasking, we observe that the forward task consistently outperforms the backward task, suggesting that predicting execution outputs provides stronger learning signals than recovering masked lines. This finding implies that output-oriented reasoning is more essential for effective code LLM training than purely reconstructive tasks.

We also find that the maximum-clique-based data curation strategy yields steady performance improvements by promoting sample diversity while reducing redundancy in the training set. 

Additionally, our incorporation of error-aware training, which explicitly leverages standard error signals, demonstrates measurable gains. It nonetheless confirms the utility of including execution errors as the informative feedback.

\begin{table}[!htb]
\centering
\setlength{\tabcolsep}{4pt}
\small
\begin{tabular}{l | cccc}
\toprule
Ablation & Total Perf.\\
\midrule
CodeBoost & \first{334.6} \\
\midrule
w/o maximum-clique cura. & 331.2 \\
w/o forward task & 330.3  \\
w/o backward task & 331.8  \\
w/o hetero. augment. & 329.0  \\
w/o errors and rewarding stderr  & 332.0  \\ % no reward stderr + no error in code
 \bottomrule
\end{tabular}
\caption{Ablation study. In each row, we exclude the specific design from CodeBoost, while keeping others as fixed. We adopt Qwen-2.5-Coder-7B-Instruct as the LLM to evaluate the effectiveness of each design, which is also used in subsequent experiments.}
\label{tab:ablation}
\end{table}

\subsubsection{Error Types.}
We next evaluate different stderr types used in training. As shown in \cref{tab:stderr}, trivially using all error types may not contribute to the highest performance. Notably, compared with syntax errors, logical errors more effectively enhance the model's ability to learn coding knowledge. The combination of both yields the optimal performance, indicating that both syntax and logical errors provide complementary and valuable training signals.

\begin{table}[!htb]
\centering
\small
\begin{tabular}{l|c}
\toprule
stderr Type &  Total Perf.\\
\midrule
all errors  &  330.5 \\
syntax errors only & 330.6 \\
logical errors only & 333.0 \\
syntax errors + logical errors & \first{334.6} \\
\bottomrule
\end{tabular}
\caption{Comparison of stderr types allowed in training code augmentation. LLMs are required to predict both stdout and stderr, even when there is no error in execution.}
\label{tab:stderr}
\end{table}

\subsubsection{Augmentation Strategy.}
Augmentation is necessary in training. We compare digit and logical augmentation strategies in \cref{tab:augmentaiton}, where the combination of both gives the highest score. Among the two strategies, digit augmentation contributes more than the logical counterpart, which indicates that more hidden knowledge can be dug out from digit perturbations.

\begin{table}[!htb]
\centering
\small
\setlength{\tabcolsep}{4pt}
% \small
\begin{tabular}{cc|ccc}
\toprule
\makecell{Digit\\Augment.} & \makecell{Logical\\Augment.} & Total Perf.  \\
\midrule
\checkmark &  & 331.5 \\
 & \checkmark & 330.0 \\
 \checkmark & \checkmark  & \first{334.6} \\
\bottomrule
\end{tabular}
\caption{Comparison of different augmentation strategies used in training.}
\label{tab:augmentaiton}
\end{table}

\subsubsection{Rewarding.}
We further evaluate the effect of replacing our rule-based reward function with an LLM-based reward model (Qwen-2.5-Coder-7B-Instruct). As shown in \cref{tab:llm_reward}, this substitution leads to a significant performance drop, suggesting that in code generation tasks where syntactic and logical correctness is critical, systematic rule-based rewards remain a more effective and reliable choice for LLM training.
\begin{table}[!htb]
\centering
\small
\setlength{\tabcolsep}{4pt}
\small
\begin{tabular}{l|ccc}
\toprule
Rewarding Type &  Total Perf.  \\
\midrule
LLM-based rewarding &  328.5 \\
rule-based rewarding (ours) &  \first{334.6} \\
\bottomrule
\end{tabular}
\caption{Comparison of the LLM-based rewarding strategy and the rule-based counterpart. We use the same LLM as the reward model (Qwen2.5-Coder-7B-Instruct).}
\label{tab:llm_reward}
\end{table}

\subsubsection{Learning Curve.}
\begin{figure}
\centering
\includegraphics[width=0.99\linewidth]{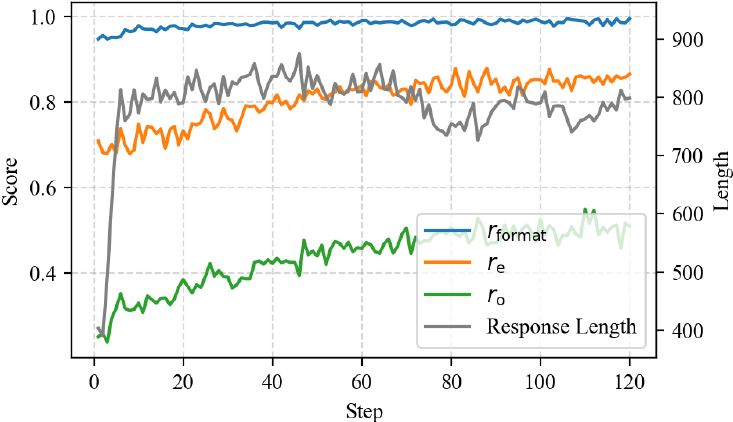}
\caption{Learning curve visualization, where reward curves and the LLM response length are plotted.}
\label{fig:plot}
\end{figure}

We subsequently plot the learning curve as shown in \cref{fig:plot}. Among the three types of rewards, the format reward $r_\mathrm{format}$ would be the easiest objective for LLMs. In contrast, the stdout reward $r_\mathbf{o}$ is shown to be the most challenging goal.
We also show the response length curve in \cref{fig:plot}, where after several steps, LLMs are using more tokens to complete the defined tasks. This behavior suggests that the models are learning to engage in more complex code reasoning processes.

\section{Limitations and Future Work}
While CodeBoost demonstrates robust performance across various code LLMs, it has certain limitations. A notable challenge lies in its current inability to effectively handle visualization-centric coding tasks. These tasks often involve interaction with graphical user interfaces (GUIs), which introduce complexities beyond standard code generation. Addressing these challenges and developing effective training strategies for code LLMs in multi-modal settings remains an important avenue for future research.

\section{Conclusion}
In this paper, we introduce CodeBoost, an RL-based post-training framework designed to enhance code LLMs without relying on human-annotated instructions. 
CodeBoost incorporates five key components to enable effective learning: 
maximum-clique curation for constructing a diverse and representative training dataset, 
bi-directional prediction to facilitate comprehensive knowledge extraction, 
error-aware prediction to leverage insights from both successful and failed executions, 
heterogeneous augmentation to enrich code semantics and diversity, 
and heterogeneous rewarding to provide fine-grained and structured supervision signals. 
Extensive experiments across multiple code LLM baselines and benchmarks validate the effectiveness of CodeBoost. 
These results underscore the scalability and potential of CodeBoost as a promising paradigm for advancing code LLMs.

%%%%%%%%%%%%%%%%%%%%%%%%%%%%%%%%%%%%%%%%%%%
%%%%%%%%%%%%%%%%%%%%%%%%%%%%%%%%%%%%%%%%%%% for supplement
\newpage
\appendix
\setcounter{figure}{0}
\setcounter{table}{0}
\setcounter{equation}{0}

\twocolumn[
\begin{center}
\vspace{1em}
\textbf{\Large Supplement}
\vspace{1em}
\end{center}
]

\renewcommand{\thesection}{S\arabic{section}}
\renewcommand{\thesubsection}{S\arabic{section}.\arabic{subsection}}
\renewcommand{\thesubsubsection}{S\arabic{section}.\arabic{subsection}.\arabic{subsection}}

\renewcommand{\thefigure}{S\arabic{figure}}
\renewcommand{\thetable}{S\arabic{table}}
\renewcommand{\theequation}{S\arabic{equation}}
%%%%%%%%%%%%%%%%%%%%%%%%%%%%%%%%%%%%%%%%%%%
%%%%%%%%%%%%%%%%%%%%%%%%%%%%%%%%%%%%%%%%%%%

\section{More Experiments}
\subsubsection{Scaling Dataset Size.}
We investigate the scalability of CodeBoost by training on subsets of varying sizes. As shown in \cref{stab:dataset_ratio}, increasing the dataset size consistently improves the overall performance. This trend suggests that our pipeline can effectively leverage larger datasets and has the potential to scale further.

\begin{table}[!htb]
\centering
\begin{tabular}{c | c}
\toprule
Dataset Ratio & Total Perf. \\
\midrule
0.25 & 330.8 \\
0.5 & 331.9 \\
1 (full) & \textbf{334.6} \\
\bottomrule
\end{tabular}
\caption{Performance comparison of using different ratios of the training dataset.}
\label{stab:dataset_ratio}
\end{table}

\subsubsection{Scaling Model Size.}
We also evaluate our CodeBoost on LLMs with varying model sizes, as presented in \cref{stab:model_size}. The results show that CodeBoost consistently enhances the performance of models across different parameter scales, including 1.5B, 3B, and 7B variants. This consistent improvement demonstrates the scalability of our method and its potential to benefit LLMs of diverse sizes.

\begin{table*}[!htb]
\centering
% \fontsize{10pt}{12pt}\selectfont % Use 9pt if necessary
\setlength{\tabcolsep}{3pt}
\small
\begin{tabular}{l|cc cc cc c | c  c c c}
\toprule
Model & \makecell{BCB (Hard)\\Complete} & \makecell{BCB (Hard)\\Instruct}  & \makecell{CRUXEval\\Output}  & \makecell{CRUXEval\\Input} & \makecell{MBPP} & \makecell{EvalPlus\\MBPP+} & \makecell{LiveCodeBench\\2501-2505} & \makecell{Total\\Perf.} \\
\midrule
Qwen2.5-Coder-1.5B-Instruct & 4.1 & 5.4 & 33.9 & 31.1 & 68.8 & 59.0 & 9.2 & 211.5\\
+ CodeBoost &  \first{6.8} & \first{6.1} & \first{36.6} & \first{32.0} & \first{69.3} & \first{60.1} & \first{10.8} & \first{221.7} \\
\midrule
Qwen2.5-Coder-3B-Instruct & 14.9 & \first{16.2} & 45.5 & 41.0 & 75.1 & 63.5 & 17.0 & 273.2  \\
+ CodeBoost & \first{16.2} & \first{16.2} & \first{48.0} & \first{43.2} & \first{76.2} & \first{63.8} & \first{18.0} & \first{281.6} \\
\midrule
Qwen2.5-Coder-7B-Instruct &  21.6 & 18.9 & 55.8 & 57.0 & 82.0 & 71.4 & 20.3 & 327.0\\
+ CodeBoost & \first{23.0} & \first{19.6} & \first{56.2} & \first{57.9} & \first{83.6} & \first{72.8} & \first{21.5} & \first{334.6} \\
\bottomrule
\end{tabular}
\caption{Performance comparison of LLMs with different sizes. After integrating with our CodeBoost, the total performance score improvements can be shown in all models. The higher scores are highlighted with bold fonts.}
\label{stab:model_size}
\end{table*}

\section{Implementation Details}
\subsubsection{Dataset Basic Filtering.}
We exclude code snippets that either fail to execute, contain visualization-related elements, or are too short (i.e., with less than 10 lines or 30 characters).

\subsubsection{Dataset Clique Curation.}
During dataset clique curation, we set the maximum size of each subset as 400, and the number of subsets $K$ is determined accordingly. We use $M=5$ iterations in the curation.

\subsubsection{Heterogeneous Augmentation.}
During heterogeneous augmentation, there usually exist errors after augmentation. We use only code snippets with supported Python's built-in error types, which include\footnote{\url{https://docs.python.org/3.10/library/exceptions.html}}: 
\begin{itemize}
\item Syntax error: SyntaxError
\item Logical errors: IndexError, ValueError, NameError, TypeError, KeyError, and ZeroDivisionError
\end{itemize}

\subsubsection{Prompt.}
\begin{figure}[!htb]
\centering
\includegraphics[width=1.05\linewidth]{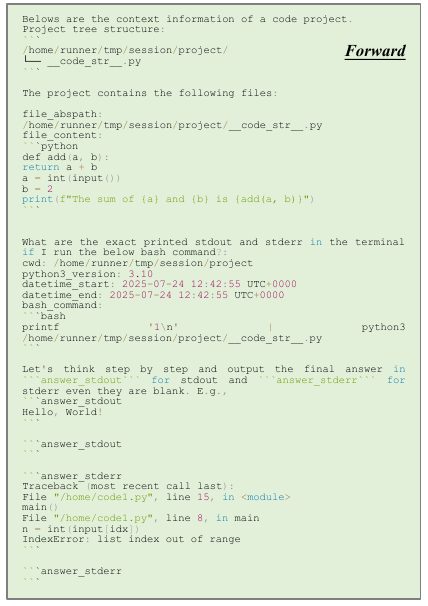}
\caption{A full prompt example for the forward task.}
\label{sfig:prompt_forward}
\end{figure}

\begin{figure}[!htb]
\centering
\includegraphics[width=1.05\linewidth]{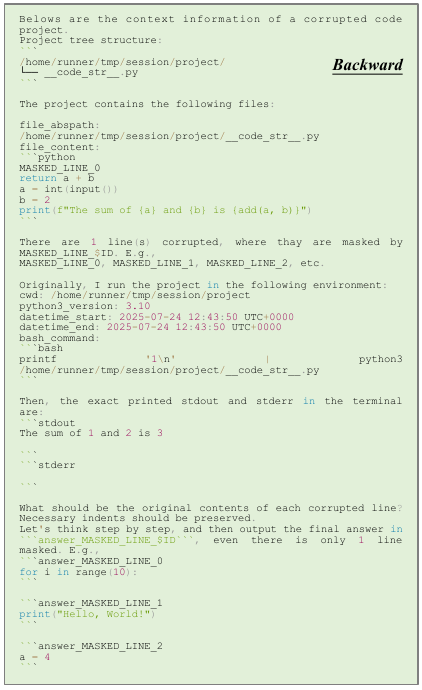}
\caption{A full prompt example for the backward task.}
\label{sfig:prompt_backward}
\end{figure}

We provide two complete prompt examples in \cref{sfig:prompt_forward,sfig:prompt_backward}. Specifically, code snippets are written into designated file paths within a project directory, whose structure is also included in the prompt. Additionally, execution date and time information are provided to help the LLM better understand the contextual environment.

\subsubsection{Code Execution.}
The code execution is achieved in a visualization-free sandbox. We set the timeout limit as 5 seconds and the RAM limit as 8 GB. Execution runs that exceed such limits will be ignored and not taken into training.

\subsubsection{Training.}
Our CodeBoost is implemented based on EasyR1, which is a modified version of verl.
In the training, we use the AdamW optimizer, with learning rate=1e-6 and weight decay=1e-2. In GRPO, we set group size=5, global batch size=128, and rollout batch size=512. 
The training for each model is in 1 epoch, which takes around 30 hours on 8 Tesla A100-80GB GPUs.

\section{Used Tools}
In this section, we list the public tools we used in our pipeline.
\subsubsection{Dataset Curation}
\begin{itemize}
\item networkx (\url{https://networkx.org/}) for extracting the maximum clique from the code snippet graph.
\end{itemize}

\subsubsection{Training}
\begin{itemize}
\item libcst (\url{https://libcst.readthedocs.io/en/latest/}) for heterogeneous augmentation.
\item swanlab (\url{https://swanlab.cn/}) for monitoring and visualizing training progress.
\end{itemize}

% \section{Datasets}
% \subsubsection{opc-sft-stage1.}
% The opc-sft-stage1~\footnote{\url{https://huggingface.co/datasets/OpenCoder-LLM/opc-sft-stage1}} dataset is 

% \subsubsection{OpenThoughts-114k.}

% \subsubsection{codeforces-cots.}

\section{Benchmarks}
\subsubsection{BigCodeBench.} 
BigCodeBench~\footnote{\url{https://github.com/bigcode-project/bigcodebench}} is a benchmark for solving practical and challenging coding tasks. 
It aims to evaluate the true coding capabilities in a realistic setting, which covers a wide variety of coding directions (such as computation, general, visualization, system, time, network, and cryptography). The benchmark is designed for HumanEval-like function-level code generation tasks, but with much more complex instructions and diverse function calls.

There are two splits in BigCodeBench. Complete: The split is designed for code completion based on the comprehensive docstrings.
Instruct: The split works for the instruction-tuned and chat models only, where the models are asked to generate a code snippet based on the natural language instructions. The instructions only contain necessary information and require more complex reasoning.

\subsubsection{CRUXEval.}
CRUXEval~\footnote{\url{https://github.com/facebookresearch/cruxeval}} is a benchmark of 800 Python functions. Each function comes
with an input-output pair. The benchmark consists of two tasks, CRUXEval-I (input prediction) and CRUXEval-O (output prediction).

\subsubsection{MBPP.}
MBPP~\footnote{\url{https://github.com/google-research/google-research/tree/master/mbpp}} consists of Python programming problems, and it is designed to be solvable by entry-level programmers, covering programming fundamentals, standard library functionality, and so on. Each problem consists of a task description, code solution, and 3 automated test cases.

\subsubsection{EvalPlus.}
EvalPlus~\footnote{\url{https://github.com/evalplus/evalplus}} is a code generation evaluation framework
to rigorously benchmark the functional correctness of LLM-generated code, which extends the test cases of the popular HumanEval and MBPP benchmarks by over 80 times.

\subsubsection{LiveCodeBench.}
LiveCodeBench~\footnote{\url{https://github.com/LiveCodeBench/LiveCodeBench}} is a challenging and contamination-free evaluation benchmark of LLMs for code that continuously collects new problems over time. LiveCodeBench annotates problems with release dates and thus allows evaluating models on problems released during a specific time period. Thus, for a newer model with a training-cutoff date, we can evaluate it on problems released after that date to measure its generalization on unseen problems.

{
    \small
    \bibliographystyle{ieeenat_fullname}
    \bibliography{ref}
}



\end{document}